# A Probabilistic Graphical Model Approach to the Structure-and-Motion Problem

Simon Streicher
Electrical and Electronic Engineering
Stellenbosch University
Stellenbosch, South Africa
sfstreicher@gmail.com

Willie Brink
Applied Mathematics
Stellenbosch University
Stellenbosch, South Africa
wbrink@sun.ac.za

Johan du Preez
Electrical and Electronic Engineering
Stellenbosch University
Stellenbosch, South Africa
dupreez@sun.ac.za

*Abstract*—We present a means of formulating and solving the well known structure-and-motion problem in computer vision with probabilistic graphical models. We model the unknown camera poses and 3D feature coordinates as well as the observed 2D projections as Gaussian random variables, using sigma point parameterizations to effectively linearize the nonlinear relationships between these variables. Those variables involved in every projection are grouped into a cluster, and we connect the clusters in a cluster graph. Loopy belief propagation is performed over this graph, in an iterative re-initialization and estimation procedure, and we find that our approach shows promise in both simulation and on real-world data. The PGM is easily extendable to include additional parameters or constraints.

## I. INTRODUCTION

The aim of this study is to consider probabilistic graphical models (PGMs) [1] for modelling and solving the well-known structure-and-motion (SaM) problem [2] in a probabilistic framework, as an alternative to the classical route of matrix factorization and the minimization of re-projection error through bundle adjustment.

PGMs provide a means to deal with uncertainty within a large system consisting of many interacting components. This is accomplished by modelling the parameters of the system as random variables, grouping the variables into clusters using the statistical dependencies between them, and connecting those clusters into a graph structure. Inference can then be performed over the graph for the determination of a posterior distribution over the variables, given observations which themselves may be characterized with uncertainty.

The goal of SaM, as used in this study, is to take a set of 2D images of a scene from various unknown viewpoints and generate 3D coordinates of feature points in the scene (the structure) as well as camera poses that can be associated with the images (the motion). The variables in such a system are the 3D feature coordinates, their observed 2D projections, and the 6-degree-of-freedom camera poses, while the mathematical relationships between these variables can be described in terms of projective geometry [3]. The SaM problem itself is not that straightforward to solve because of the inter-dependency between structure and motion, the fact that absolute scale is lost in projection, and the inherently ambiguous nature of image data.

We develop and investigate a fairly general technique for parameterizing systems with continuous random variables and nonlinear relationships between them. The idea is to approximate certain distributions with Gaussians, for their mathematical convenience and familiarity within the PGM literature. We will use the unscented transform to linearize nonlinear relationships between variables, and combine the variables into clusters. Observations can then be used to update and propagate beliefs through the graphical model, in order to find a posterior distribution over the latent variables.

Formulating systems in terms of PGMs enables immediate expansion to include additional logic, such as to penalize conflicting parameters, as well as integration with other probabilistic models to form complex systems. A probabilistic approach also provides additional insight into the system, such as a confidence over estimated parameter values.

We start with a brief overview of projection geometry, and then explain our PGM formulation along with the manner in which we linearize and an overview of our implementation. Experimental results are given and discussed, and we end the paper with some concluding remarks.

## II. PROJECTIVE GEOMETRY

The mathematical relationships between 3D feature coordinates, their 2D projections in the respective images, and the camera poses associated with the images, can be described elegantly in homogeneous coordinates [3] (which form the basis of projective geometry).

The pinhole camera model, depicted in Fig. 1, takes a 3D point $\mathbf{X}$ in world coordinates to a 2D point $\mathbf{x}$ on the image plane of a particular camera. As the name of the model suggests, this projection is done through a single point $\mathbf{C}$ in world coordinates which we call the optical centre of the camera. The action of a pinhole camera can be expressed very succinctly in homogeneous coordinates as

$$\mathbf{x} = \mathbf{PX}, \tag{1}$$

where $\mathbf{X}$ is a point in $\mathbb{P}^3$, P a $3\times 4$ homogeneous matrix, and $\mathbf{x}$ a point in $\mathbb{P}^2$. The equality sign here indicates homogeneous equivalence, implying that the vectors are equal only up to scale.

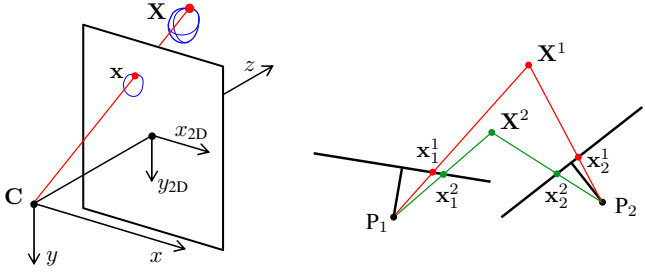

Fig. 1. The pinhole camera model takes a 3D feature point $\mathbf{X}$ and projects it in a straight line through the optical centre $\mathbf{C}$ to a point $\mathbf{x}$ on the 2D image plane. It is often convenient to consider this situation in one dimension lower, as in the diagramme on the right. The projected coordinates of a feature in two or more spatially separate image planes can be used to recover its original 3D coordinates, if the camera poses are known.

The matrix P in equation (1) is called the camera projection matrix, or camera matrix for short, and contains the parameters of the camera in the form

$$\mathrm{P} = \mathrm{KR}[\mathbf{I} \,|\, -\widetilde{\mathbf{C}}]. \qquad (2)$$

Here I denotes the $3 \times 3$ identity matrix, the vector $\widetilde{\mathbf{C}}$ is the Euclidean version of the camera centre in world coordinates, and the rotation matrix R describes the rotation from world coordinates to camera coordinates (the $xyz$-axes shown in Fig. 1). R and $\mathbf{C}$ are the extrinsic parameters, and describe the position and orientation (i.e. the pose) of the camera in world coordinates. The $3 \times 3$ calibration matrix K contains the intrinsic parameters of the camera, like its focal length and principal point offsets, to scale and translate camera coordinates to image coordinates in pixels.

Consider a set of $m$ images, each taken with a camera at some distinct position and orientation in world coordinates, and $n$ features identified across these images. Let $\mathbf{x}_j^i$ be the homogeneous image coordinates of feature $i$ in image $j$, that is

$$\mathbf{x}_j^i = \mathrm{P}_j \mathbf{X}^i. \qquad (3)$$

The structure-and-motion (SaM) problem asks for both camera matrices $\mathrm{P}_j$, $j = 1, 2, \ldots, m$ and 3D feature coordinates $\mathbf{X}^i$, $i = 1, 2, \ldots, n$, given these image coordinates. For the purposes of this paper we assume knowledge of calibration matrices, so the unknowns in $\mathrm{P}_j$ are only the extrinsic parameters of camera $j$.

Identifying and matching the projections of features across a set of images can be accomplished through a method like SIFT [4] or ORB [5], and outliers can be discarded by enforcing geometric constraints in a RANSAC process [6]. A solution to the SaM problem can then be attempted through repeated pairwise estimation of camera poses and feature point triangulations [7], which involve matrix decomposition. The issue of drift can be addressed through bundle adjustment [8], where camera poses and feature points are altered in order to minimize re-projection error.

With this study we offer an alternative formulation of the SaM problem that allows for a solution to be achieved through probabilistic reasoning.

## III. OUR PGM FORMULATION

The general idea of solving the SaM problem with a PGM is to use prior information about the unknown camera poses and 3D feature coordinates, and integrate that with observed 2D projections in order to obtain posterior knowledge about the situation. We accomplish this by

- specifying the parameters of the system as random variables,
- specifying the relationships between these variables as well as prior distributions over them,
- categorizing the variables into clusters,
- constructing a cluster graph from the clusters,
- and finally running belief propagation over the cluster graph to obtain a posterior distribution.

These steps result in a posterior distribution over the variables, and can be viewed as a fairly general approach to probabilistic modelling. If needed, a solution to the problem can be taken as those values with highest probability. Importantly though, this approach gives a distribution, or measure of uncertainty over the solution.

### A. Random variables

The random variables in a SaM system are the pose variables for every camera, the 3D coordinates for every feature seen in two or more images, and the image coordinates for every feature in every image that it is seen.

Every camera matrix $\mathrm{P}_j$ is converted to a vector containing the Euclidean coordinates of the camera centre as well as the Euler angles representation of its rotation matrix, i.e. something of the form

$$\mathbf{p} = [C_x,\ C_y,\ C_z,\ \theta_x,\ \theta_y,\ \theta_z]^T. \qquad (4)$$

We work with Euclidean versions of $\mathbf{X}^i$ and $\mathbf{x}_j^i$, since the arbitrary scale in a homogeneous vector would complicate the meaning of a probability distribution. For simplicity and neatness we redefine the symbols $\mathbf{X}^i$ and $\mathbf{x}_j^i$ in the rest of this paper to refer to 3D feature coordinates and 2D image coordinates, respectively, and introduce $\mathbf{p}_j$ to denote the vector representation of the extrinsic parameters of camera $j$.

From now on we view the variables $\mathbf{X}^i$, $\mathbf{p}_j$ and $\mathbf{x}_j^i$ as random variables (of dimension 3, 6 and 2 respectively). An observation will be indicated with a hat, for example $\widehat{\mathbf{x}}_j^i$. In essence, the SaM problem can be formulated as finding a posterior over $\mathbf{X}^1, \ldots, \mathbf{X}^n$ and $\mathbf{p}_1, \ldots, \mathbf{p}_m$ given the observed projections $\widehat{\mathbf{x}}_j^i$.

### B. Relationships between variables

In order to relate observations to latent variables, we need relationships between them. In our case we may define a function $f$ such that

$$\mathbf{x}_j^i = f(\mathbf{p}_j, \mathbf{X}^i). \qquad (5)$$

That is to say, $f$ constructs a camera matrix $\mathrm{P}_j$ from the elements of $\mathbf{p}_j$ (which is a nonlinear operation on the angles), converts $\mathbf{X}^i$ from $\mathbb{R}^3$ to $\mathbb{P}^3$, then applies the projection stated in equation (3), and finally converts the result from $\mathbb{P}^2$ to $\mathbb{R}^2$ (which is also nonlinear).

## C. Building a cluster graph

As mentioned, we use cluster graphs to model the SaM problem. An alternative would be to consider factor graphs. It seems as if the latter is more widely used, possibly due to the fact that they are trivial to construct (while the construction of a valid cluster graph often requires careful planning). A comparative study found belief propagation to converge quicker over cluster graphs, with seemingly the same accuracy [9]. This may be attributed to the fact that the multivariate sepsets in cluster graphs allow for information between variables to be propagated more effectively.

Figure 2 (top) shows a small SaM example represented as a Bayesian network to demonstrate the dependencies between variables. From such a network we can extract all the factors in the joint over the variables, which would be the priors $p(\mathbf{X}^i)$ and $p(\mathbf{p}_j)$ and the conditionals $p(\mathbf{x}_j^i \mid \mathbf{p}_j, \mathbf{X}^i)$. A cluster can be defined for every factor, so we group the parameters involved in a single projection into a cluster $C_j^i = \{\mathbf{x}_j^i, \mathbf{p}_j, \mathbf{X}^i\}$.

A cluster graph is an undirected graph with clusters as nodes, and connections between clusters via sepsets. A sepset between two clusters must be contained in the intersection of those two clusters, and all sepsets must abide by the running intersection property. This property states that for any variable $x$ in the graph, any two clusters containing $x$ must have a unique path of sepsets containing $x$ between them [1].

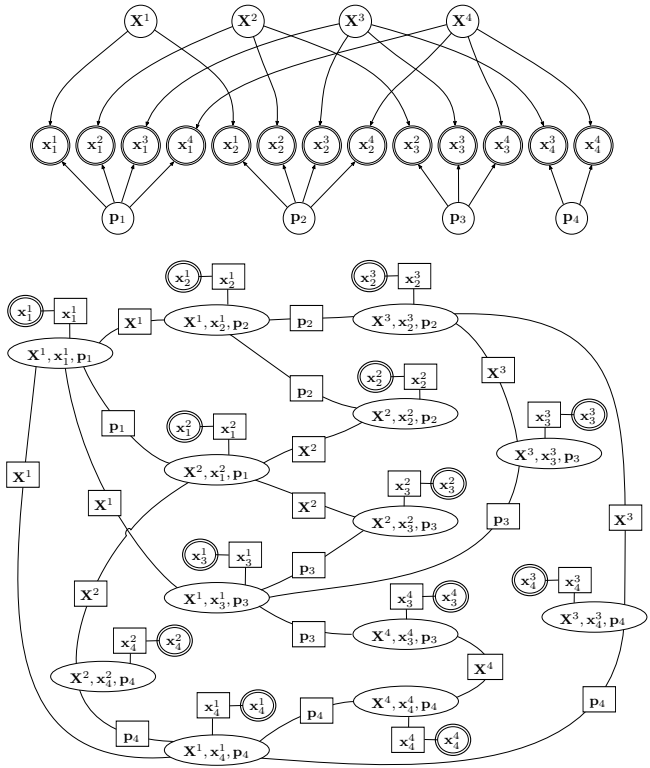

Fig. 2. A small example of a structure-and-motion problem involving four cameras and four features, drawn as a Bayesian network (top) and a cluster graph (bottom). Double-circles indicate observed variables, and note that features are not necessarily visible to all cameras.

In this study we make use of Du Preez's algorithm to build cluster graphs, as detailed in [10]. The algorithm formulates sepsets by iterating through the random variables and connecting every cluster containing a particular variable in a tree structure, before superimposing all of these trees. The result for our small example is shown in Fig. 2 (bottom).

## D. Belief propagation

The next step would be to consider the observed variables and pass that information and its implications through the cluster graph in order to reach a consensus.

Since features must be observed in at least two cameras for their 3D coordinates to be estimated, our graph will contain loops. This calls for loopy belief propagation [1], which is an iterative message passing scheme. Although it is not proven that this approach will converge or that the posterior obtained will be a good representation of the true distribution, it can still be of practical use if implemented effectively.

The random variables in our system are continuous. A significant problem now stems from the nonlinearities in the relationships between them, specifically in the function $f$ in equation (5), as well as the operations performed on distributions during belief propagation. It is near impossible to define arbitrary continuous distribution functions in closed-form, and expect the result of a belief update to be in the same parametric family. Thus some form of approximation becomes necessary. A Monte-Carlo approach might seem possible, but the size of a typical SaM problem would require an exorbitant number of samples.

For mathematical convenience and practical feasibility we decide to model every observed variable, which is a projection $\mathbf{x}_j^i$, as a Gaussian random variable with mean the measured coordinates and some standard deviation in pixels caused by measurement noise. We then take steps to force other distributions in the system to also be Gaussian, thereby effectively linearizing the nonlinear operations brought about by projection and message passing.

## IV. Linearization

We choose to implement the unscented transform for linearizing nonlinear variable transformations and maintaining Gaussian distributions during the process of message passing.

### A. The unscented transform

Consider a Gaussian distribution with mean $\boldsymbol{\mu}$ and covariance $\Sigma$. With sufficiently many measurements generated from this distribution, the sample mean $\widehat{\boldsymbol{\mu}}$ and sample covariance $\widehat{\Sigma}$ ought to reflect the parameters of the original distribution closely. In fact, a minimal set of weighted points (called sigma points) can be contrived in such a way as to hold the exact relationships $\widehat{\boldsymbol{\mu}} = \boldsymbol{\mu}$ and $\widehat{\Sigma} = \Sigma$. Furthermore, extracting a set of sigma points from a linearly transformed Gaussian distribution is equivalent to simply transforming the sigma points of the original distribution.

The process of changing the parametric representation of a Gaussian distribution to a set of sigma points is referred to as the unscented transform [11]. Among many choices, the following two formulations can be shown to be valid [12]. From an $d$-dimensional Gaussian distribution we may define $2d$ points and corresponding weights as follows:

$$\mathbf{s}_{\pm i} = \boldsymbol{\mu} \pm k\boldsymbol{\ell}_i \text{ and } w_i = 1, \quad i = 1, \ldots, d, \quad (6)$$

where $\boldsymbol{\ell}_i$ is the $i$th column of the lower-triangular matrix in the Cholesky factorization of $\Sigma$. Alternatively, the so-called standard sigma point formulation defines $2d+1$ points as

$$\mathbf{s}_0 = \boldsymbol{\mu}, \ \mathbf{s}_{\pm i} = \boldsymbol{\mu} \pm \sqrt{\frac{d}{1-w_0}}\boldsymbol{\ell}_i, \quad i = 1, \ldots, d, \quad (7)$$

with weights

$$0 < w_0 < 1, \ w_{\pm i} = \frac{1-w_0}{2d}, \quad i = 1, \ldots, d. \quad (8)$$

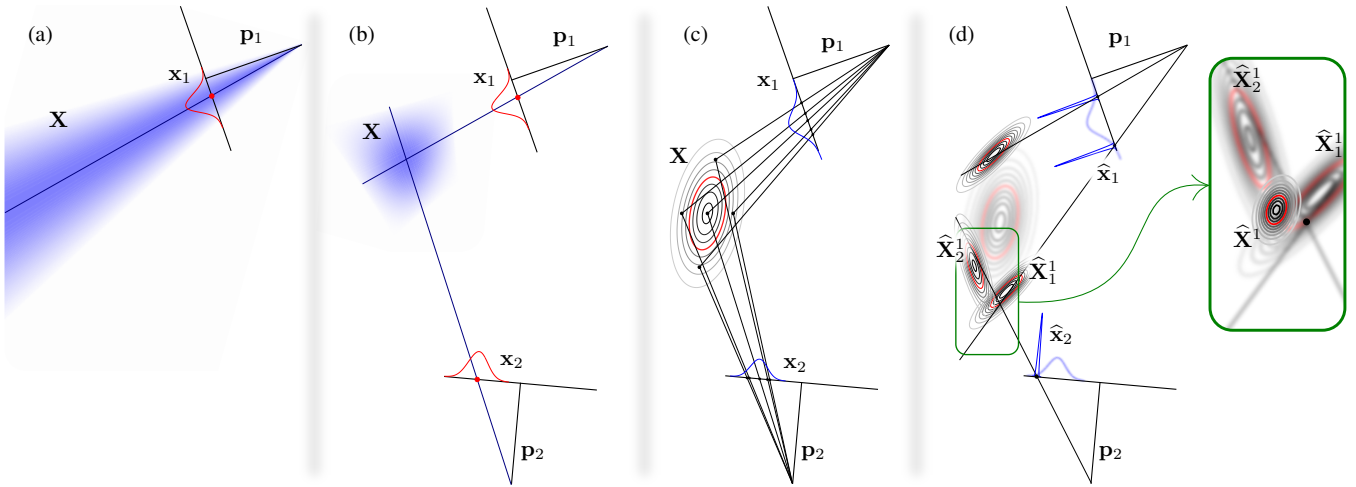

Fig. 3. An illustration of our probabilistic analogy to stereo view triangulation from known camera poses, using sigma point parameterization. (a) The back-projection of a Gaussian distribution on the image plane gives a non-Gaussian distribution over the 3D feature $\mathbf{X}$. (b) A second projection from a different viewpoint results in a distribution over $\mathbf{X}$ that is closer to being Gaussian. (c) By projecting the sigma points of the prior over $\mathbf{X}$, we get distributions over $\mathbf{x}_1$ and $\mathbf{x}_2$. (d) Observation of these projections then yields the posterior over $\mathbf{X}$, as seen in the insert on the right.

The unscented transform can be particularly useful in approximating a nonlinearly transformed Gaussian distribution with a Gaussian distribution. To this end, consider a Gaussian random variable $\mathbf{x}$ and some nonlinear transformation $\mathbf{y} = f(\mathbf{x})$. Since $f$ is nonlinear, the variable $\mathbf{y}$ is not Gaussian (and, in the context of belief propagation, may be quite difficult to describe). We can take a set of sigma points $\{\mathbf{s}_i\}$ calculated from $\mathbf{x}$, transform each to $\mathbf{t}_i = f(\mathbf{s}_i)$, and approximate $\mathbf{y}$ as a Gaussian with mean and covariance equal to the mean and covariance of the transformed set $\{\mathbf{t}_i\}$.

This linearization with sigma points is simple and efficient to implement and, unlike a Taylor linearization, does not need the Jacobian of the transformation function. It also captures the mean and covariance accurately to the third-order Taylor expansion for any nonlinearity [12].

Sigma points are also useful for estimating the joint $p(\mathbf{x}, \mathbf{y})$ as a Gaussian distribution. The sigma points of $\mathbf{x}$ and $\mathbf{y}$ are simply concatenated as $[\mathbf{s}_i^T, \mathbf{t}_i^T]^T$, to form a new set of sigma points which parameterizes the joint distribution.

### B. Stereo triangulation example

Consider the situation in Fig. 3(a), where the projection of a feature with unknown coordinates $\mathbf{X}$ is measured to be $\widehat{\mathbf{x}}_1$ on the image plane of a camera parameterized by $\mathbf{p}_1$. The projection can be described with a Gaussian random variable $\mathbf{x}_1$, but it is clear that back-projection results in a distribution over $\mathbf{X}$ that is very much non-Gaussian. If a second projection is considered, as in Fig. 3(b), the distribution over $\mathbf{X}$ becomes closer to a Gaussian, and we can use sigma points to approximate the distribution with a Gaussian.

Now consider the scenario in Fig. 3(c) where $\mathbf{X}$ is modelled as Gaussian, with some prior distribution (of which a few contours are shown in the figure). If the camera parameters are known, the sigma points determined from $\mathbf{X}$ can be projected to the two image planes to produce Gaussian distributions over $\mathbf{x}_1$ and $\mathbf{x}_2$. A posterior distribution over $\mathbf{X}$ can now be found by observing $\widehat{\mathbf{x}}_1$ and $\widehat{\mathbf{x}}_2$, as indicated in Fig. 3(d) (for illustration the figure shows two different observations of $\mathbf{x}_1$; the one marked $\widehat{\mathbf{x}}_1$ results in the posterior over $\mathbf{X}$ shown in the insert). This posterior is the combined result of the prior and the two observations, and the procedure is a probabilistic analogy to stereo view triangulation.

### C. Full sigma point parameterization

Prior distributions over the latent variables in our system (the 3D feature coordinates $\mathbf{X}^i$ and camera poses $\mathbf{p}_j$) can be turned into a prior for every cluster $C_j^i = \{\mathbf{x}_j^i, \mathbf{p}_j, \mathbf{X}^i\}$ in our cluster graph. We accomplish this by sigma point parameterizing the priors over $\mathbf{X}^i$ and $\mathbf{p}_j$, and applying the projection in equation (5) to every combination, as illustrated in Fig. 4. This yields a set of sigma points over $\mathbf{x}_j^i$, which we combine with those of $\mathbf{X}^i$ and $\mathbf{p}_j$ to obtain a joint over all three variables. Observations of the form $\widehat{\mathbf{x}}_j^i$ can now be propagated through the cluster graph, for updated beliefs over $\mathbf{X}^i$ and $\mathbf{p}_j$.

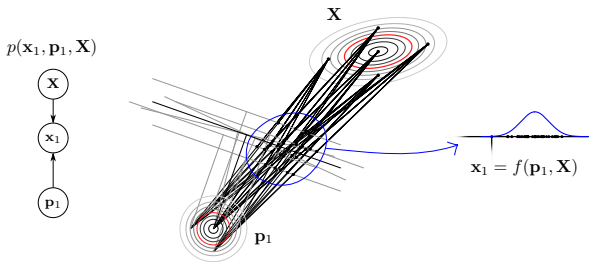

Fig. 4. Every combination of sigma points parameterizing prior distributions over $\mathbf{X}$ and $\mathbf{p}$ can be used in a projection to produce a sigma point parameterization of $\mathbf{x}$ as shown on the right.

## V. IMPLEMENTATION

We now summarize how a structure-and-motion problem may be solved by our PGM approach. The input is assumed to be image feature correspondences identified across a set of images. We set up clusters corresponding to every projection, and construct a cluster graph. Gaussian priors are placed on all 3D feature coordinates and camera poses, and transferred to the projections. We observe the projections with a small amount of uncertainty, and loopy belief propagation is performed over the cluster graph, using Gaussian factors in the manner prescribed by Koller and Friedman [1], to arrive at a joint posterior over the latent variables.

Since the linear approximations are derived from priors, the system models the space in the vicinity of these priors more accurately than the space further away. Therefore, if the priors do not encompass the true solution, the linearized projective transformations may become unstable. To circumvent this problem we allow the system to recapture the projective geometry a posteriori, by repeating the following:

- re-initialize new sigma points for every cluster with the means and covariances of the current posterior distribution,
- run belief propagation on the new clusters to obtain yet a new posterior.

A further complication may arise here, namely that the covariances of the clusters are likely to get smaller for every new posterior. As a result, the system may converge to a local minimum far from the true solution. We remedy this by enlarging the covariances to allow the system to escape and explore the space outside a local minimum.

## VI. EXPERIMENTAL RESULTS

We proceed with some experimental results. Firstly, we demonstrate our approach in one dimension lower. Refer to Fig. 5. We randomly position 15 features in 2D, and place 7 cameras randomly around these points. The features are projected to the 1D image plane of every camera, and some of these projections are dropped to simulate the idea that not all features are observed by all cameras. The remaining projections serve as our observations. As a prior for every 2D point we choose a wide Gaussian centred at a fixed point in the middle of the scene, and as priors for the cameras we centre

(a) initial camera poses and feature positions, with ground truth camera poses in light grey, and a graph indicating re-projection errors (coloured according to cameras)

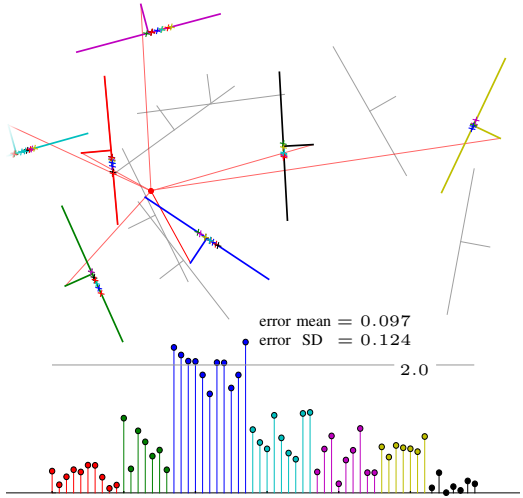

(b) the system after a local minimum has been reached

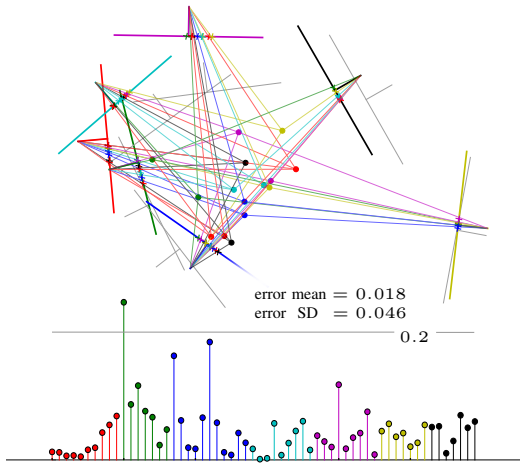

(c) the system after the final position has been reached

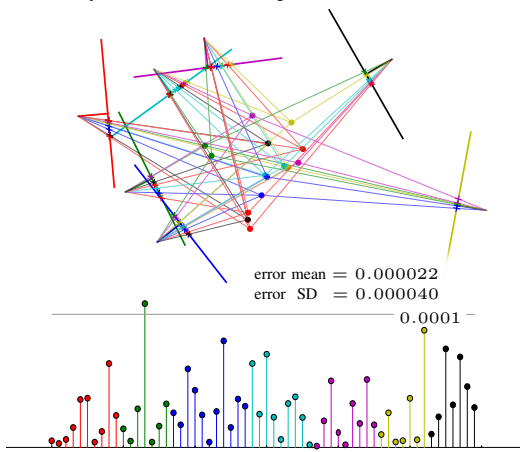

Fig. 5. An example of solving a synthetic 2D structure-and-motion problem. Every plot shows the ground truth camera positions in light grey, and note that the plots are anchored to the red camera's pose (to avoid the similarity ambiguity in the solution).

TABLE I
THE EFFECT OF NOISY PRIORS ON THE AVERAGE RE-PROJECTION ERROR,
PRIOR TO AND AFTER BELIEF PROPAGATION (TOP AND BOTTOM NUMBER
IN A CELL, RESPECTIVELY), FOR VARIOUS NUMBERS OF CAMERAS,
NUMBERS OF FEATURES AND EXPECTED MEASUREMENT NOISE.

| | | | standard deviation of added noise | | | |
|---|---|---|---|---|---|---|
| | | angles → | 2.5° | 5.0° | 10.0° | 20.0° |
| | | position → | 0.5 | 1.0 | 2.0 | 4.0 |
| cams ↓ | feats ↓ | $\sigma$ ↓ | prior and posterior re-projection error | | | |
| 5 | 50 | 0.0001 | 0.1433 / 0.0002 | 0.2356 / 0.0005 | 0.3811 / 0.0004 | 3.1203 / 0.0020 |
| 5 | 100 | 0.0007 | 0.1495 / 0.0010 | 0.2174 / 0.0016 | 0.2366 / 0.0029 | 0.5091 / 0.0060 |
| 7 | 60 | 0.0004 | 0.1276 / 0.0004 | 0.1993 / 0.0010 | 0.2438 / 0.0026 | 0.7232 / 0.0036 |
| 10 | 100 | 0.0012 | 0.1419 / 0.0012 | 0.2156 / 0.0018 | 0.3697 / 0.0066 | 1.0230 / 0.0093 |
| 10 | 200 | 0.0020 | 0.1382 / 0.0014 | 0.2236 / 0.0031 | 0.4711 / 0.0079 | 0.8315 / 0.0114 |
| 20 | 200 | 0.0030 | 0.1471 / 0.0020 | 0.1998 / 0.0041 | 0.4000 / 0.0054 | 0.7868 / 0.0180 |
| 30 | 500 | 0.0040 | 0.1479 / 0.0030 | 0.1972 / 0.0051 | 0.3212 / 0.0100 | 0.8402 / 0.0203 |

Gaussians around significant perturbations of the ground truth poses. Figure 5 depicts the prior situation in (a), the result of one pass of loopy belief propagation in (b), and the final optimal solution obtained from our iterative re-initialization scheme in (c). The system converged to the correct solution, even though initial estimates were quite far off.

We also simulated the SaM problem in full 3D. Features are generated randomly inside a sphere of radius 2, and cameras are placed randomly on a sphere of radius 10 so that they face the centre. Gaussian noise is added to the camera parameters. Input 2D feature coordinates are found by projecting the 3D features with the respective camera parameters. We then add noise to the camera poses and 3D feature coordinates before we fit priors over them, while keeping the projections fixed. Table I lists the average re-projection errors for increasing amounts of noise in the priors, and we see clear improvements after running belief propagation. The outcome suggests that the accuracy of the posterior distribution is affected by the level of noise in the prior, which is to be expected. The table also shows the effect of the expected measurement noise, which is the standard deviation $\sigma$ in the Gaussians centred around measured image feature coordinates.

Finally, as a proof of concept, we demonstrate our technique on a real-world example. We take a set of images, extract and match features, and follow the classical matrix factorization route to arrive at an initial configuration. We then centre priors with wide variance around this initial solution, and perform belief propagation over the associated cluster graph for refinement. The result is shown in Fig. 6.

## VII. CONCLUSION

The objective of this study was to investigate a probabilistic approach to model and solve the structure-and-motion problem which involves continuous variables and nonlinearities. We made use of sigma point parameterizations to capture the links between variables as covariances of Gaussian distributions. We found it best to perform an iterative re-initialization and loopy-belief procedure, with widened covariances to escape local optima (an approach not unlike simulated annealing). Experimentation suggests that there is promise in this approach.

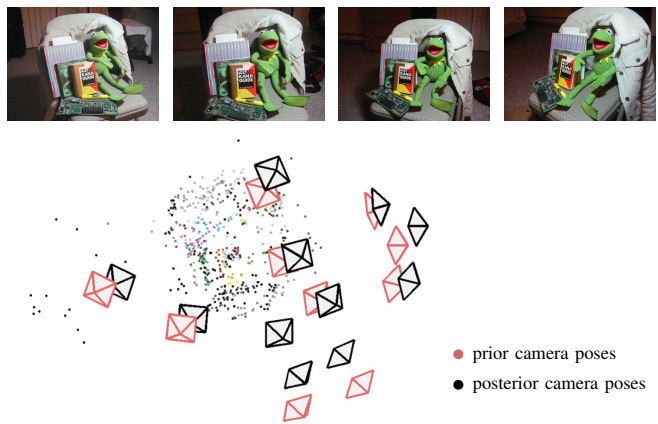

Fig. 6. A real-world structure-and-motion problem solved by our PGM approach. Priors were created from wide Gaussians around an initial solution (found by matrix factorization). The 3D plot is anchored to the first camera pose, and only 4 of the 11 input images are shown.

Our PGM can be extended to include additional parameters, such as the intrinsic camera parameters or binary random variables to allow the system to infer the correctness of feature matches (which is an unresolved issue in most structure-and-motion systems).